Journal on
Artificial Intelligence

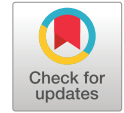

Tech Science Press

# Modeling & Evaluating the Performance of Convolutional Neural Networks for Classifying Steel Surface Defects

**Nadeem Jabbar Chaudhry[1,*], M. Bilal Khan[2], M. Javaid Iqbal[1] and Siddiqui Muhammad Yasir[3]**

[1]Department of Computer Science, Superior University, Lahore, Pakistan
[2]Department of Computer Science, University of Central Punjab, Lahore, Pakistan
[3]Department of Mechanical System Engineering, Tongmyong University, Busan, Korea
*Corresponding Author: Nadeem Jabbar Chaudhry. Email: nadeem.ch@superior.edu.pk

**Abstract:** Recently, outstanding identification rates in image classification tasks were achieved by convolutional neural networks (CNNs). to use such skills, selective CNNs trained on a dataset of well-known images of metal surface defects captured with an RGB camera. Defects must be detected early to take timely corrective action due to production concerns. For image classification up till now, a model-based method has been utilized, which indicated the predicted reflection characteristics of surface defects in comparison to flaw-free surfaces. The problem of detecting steel surface defects has grown in importance as a result of the vast range of steel applications in end-product sectors such as automobiles, households, construction, etc. The manual processes for detections are time-consuming, labor-intensive, and expensive. Different strategies have been used to automate manual processes, but CNN models have proven to be the most effective rather than image processing and machine learning techniques. By using different CNN models with fine-tuning, easily compare their performance and select the best-performing model for the same kinds of tasks. However, it is important that using different CNN models either from fine tuning can be computationally expensive and time-consuming. Therefore, our study helps the upcoming researchers to choose the CNN without considering the issues of model complexity, performance, and computational resources. In this article, the performance of various CNN models like Visual Geometry Group, VGG16, VGG19, ResNet152, ResNet152V2, Xception, InceptionV3, InceptionResNetV2, NASNetLarge, MobileNetV2, and DenseNet201 with transfer learning techniques are evaluated. These models were chosen based on their popularity and impact in the field of computer vision research, as well as their performance on benchmark datasets. According to the outcomes, DenseNet201 outperformed the other CNN models and had the greatest detection rate on the NEU dataset, falling in at 98.37 percent.

**Keywords:** Convolutional neural network; transfer learning; defect detection

## 1 Introduction

The tensile strength of the steel manufacturing business plays a significant role in the manufacturing industry. Steel is widely used in a variety of industries, including construction, infrastructure, trains, automobiles, bridges, machines, ships, household materials, tools, etc. Steel quality has a direct impact on the final result of the end product. Poor-quality steel might result in decreased strength, higher friction, and a poor look. So, to check the quality of produced steel, a variety of reasonable manual methods like an inspection by quality control personnel are used. These manual processes are labor-intensive, expensive, and frustrating. The major problems in manual processes [1] are as follows:

- The steel manufacturing companies are continuously 24 h in operational so quality control staff have to work at awkward timing like late at night. Which sooner or later creates a negative impact on defect detection and worker health.
- The quality control personnel have to continuously look at the steel plates. It eventually becomes the cause of mental and visual tiredness and fatigue, turning to leads error [2,3].
- The quality control personnel have a very shortage. Time to identify the defeated piece of steel due to continuity in rolling steel strips. This may become the cause of overlooking the defective steel.

The automation for the classification of the steel surface is not just for quality improvement but also used for speeding up the entire manufacturing process. Automation will help in; Consistency: Automated systems can perform the same task repeatedly without getting tired or making mistakes, ensuring consistent output. Efficiency: Automated systems can complete tasks faster and with higher accuracy than humans, leading to increased productivity and reduced costs. Scalability: Automated systems can easily scale up or down depending on the workload, eliminating the need for hiring additional staff. Cost-effectiveness: Automated systems can be more cost-effective in the long run than hiring and training human workers.

Computer vision techniques are frequently employed to automate industrial problems and have become a quality control flag mark. In recent years, Fig. 1 convolution neural network (CNN) based approaches and techniques have not only produced outstanding results in the classification of a large number of objects [4–6] due to their entirely automatic solutions [7–10] but have also performed well in the classification of surface defects [11].

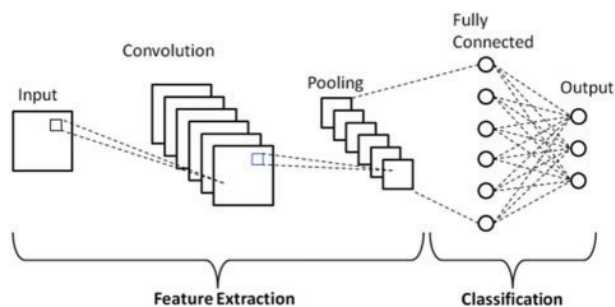

**Figure 1:** Basic architecture of convolutional neural network (CNN) [12]

The results of the steel classification with CNN [13,14] show that CNN Fig. 1 base techniques perform well rather than the image processing-based methods and machine learning approaches. AlexNet, a CNN model, was the first to be introduced for large-size datasets and performed well [6], followed by VGG [15] and ResNet [16], which also improved the results on huge classes of the dataset. The surface defect of Fabric [17–19], semiconductor [20], timber [21], and asphalt pavement [22,23] are solved using different CNN models as the backbone. They have the same kind of issue as steel defects classification.

This work aims to do a systematic study of transfer learning effectiveness for steel surface defects classification. Transfer learning aims to reuse the learned feature of one domain for the same kind of other domain [24] for reducing the computation cost and resources. By using different CNN models with fine-tuning, easily compare their performance and select the best-performing model for such kinds of tasks. This approach can also help to identify the most important features and patterns in the data, as different CNN models may learn different representations of the data. However, it is important that using different CNN models with fine tuning can be computationally expensive and time-consuming. It also required a large amount of labeled data to train and fine-tune each model. Therefore, our study helps the upcoming researchers to choose the CNN without considering the issues of model complexity, performance, and computational resources when using this approach. The outcome of this article can be further improved in the future by fine-tuning and designing a new deep architecture for steel surface defects classification.

This paper is organized as follows: Section 2 is about the literature review. Section 3 describes the used methodology in the article with some details of transfer learning and used CNN models. Section 4 discusses the dataset, pre-processing strategies, and the results. The last, Section 5, is describing the conclusion of the article.

## 2  Related Work

Zhou et al. [25] created their own CNN model and applied it to a custom basis dataset. They were able to lower the error rate to 0.6292 percent. Fu et al. [26] suggested a multi-respective field module that achieved up to 100 percent accuracy on the benchmark NEU dataset using SqueezeNet as a base model. He et al. [27], suggested a hierarchical learning architecture based on CNN that achieved up to 97.2 percent accuracy using two separate custom-built datasets. The majority voting mechanism combined with CNN was utilized to detect hot-rolled steel strip datasets with 99.50 percent accuracy [28]. He et al. [16], the author employed a multi-group CNN model to achieve 94 percent accuracy on a dataset of hot-rolled steel strips. Masci et al. [29] purposed a multi-scale pyramidal pooling network approach based on CNN for defect classification of the steel surface. That approach can also perform on non-equal image sizes; this approach can be seen as a fully supervised hierarchical bag-of-features extension.

Di et al. [30] suggested a semi-supervised generative adversarial network (GAN) for the classification of a dataset of hot-rolled steel strips and achieved good results with up to 98.2% accuracy. Hu et al. [31] suggested a genetic approach using a hybrid chromosome and a support vector machine (SVM) for classification on a dataset of hot-rolled steel strips, with a 95.04 percent accuracy. Liu et al. [32], used SVM and the shearlet transform to reach an accuracy of 97.71 percent. On the hot-rolled steel, Gong et al. [33] developed an adjustable hyper-sphere with SVM and achieved an accuracy of up to 93.08 percent. Chu et al. [34] author used quantile hyper-spheres and SVM to achieve 96.16 percent accuracy on a custom-built dataset. SVM and hyper-spheres with noise insensitivity were also used by Gong et al. [33] On the hot-rolled steel strip dataset, the accuracy was 97.18 percent. Zhao et al. [35] used the distance of the local descriptor with manifold to construct the linear models. They achieved up to 87.36% accuracy on Strip Steel Defect Dataset. Furthermore, Ongoing methods for detecting steel defects are thoroughly explained in Table 1.

Ma et al. [36] used local and global features with neighborhood gray structures to achieve significant accuracy in detecting surface defects. Wang et al. [37] purposed a complex calculation-based method through a histogram of oriented gradient and gray level co-occurrence matrix, but that approach was noise sensitive. Bulnes et al. [38] purposed both the classification and detection of defects for steel surfaces and achieved 96.52 and 80.30 percent accuracy respectively. The detection was based on the division of the images into a set of overlapping areas and the optimum value for defect detection was automatically generated through a genetic algorithm. The neural network was used for classification.

On the one hand, CNN models produce the best accuracy but at the same time training a model consumes a high number of resources and time. To beat these issues, transfer learning is the suggested strategy. Transfer learning is a machine learning technique throughwhich a pre-trained model on a large dataset will use for the same kind of other problem. Fine-tuning can be applied for enhancing the results.

**Table 1:** An overview of recent steel surface defect classification techniques

| Year | Author | Dataset | Results | Classifier |
|---|---|---|---|---|
| Zhou et al. (2017) | [25] | Hot-rolled steel sheet | 0.62% error rate | Custom base CNN |
| Fu et al. (2019) | [26] | NEU | 100% | SqueezeNet + multi-respective field module |
| He et al. (2019) | [27] | Custom build | 97.2% | Hierarchical learning framework + CNN |
| Natarajan et al. (2017) | [28] | Hot-rolled steel strip | 75.26% | CNN with the mechanism of majority voting |
| He et al. (2019) | [16] | Hot-rolled steel strip | 96% | Multi-group CNN |
| Masci et al. (2013) | [29] | Columbia-Utrecht | 75.3% | CNN of multi-scale pyramidal pooling network |
| Di et al. (2019) | [30] | Hot-rolled steel strip | 98.8% | Semi-supervised GAN |
| Hu et al. (2016) | [31] | Hot-Rolled steel strip | 95.04% | SVM + hybrid chromosome genetic algorithm |
| Liu et al. (2020) | [32] | Custom build | 97.71% | SVM + shearlet transform |
| Chu et al. (2017) | [34] | Hot-rolled steel strip | 96.16% | Adjustable hyper-sphere + SVM |
| Gong et al. (2019) | [33] | Hot-rolled steel strip | 97.26% | Hyper-spheres with insensitivity to noise + SVM |
| Zhao et al. (2018) | [35] | Strip steel defect dataset | 99.82% | The nearest neighbor classifier of the local descriptor |
| Ma et al. (2017) | [36] | Custom build | 100% | Local and global features with neighborhood gray structure |
| Wang et al. (2018) | [37] | Custom build | 91% | Histogram of oriented gradient and gray level co-occurrence matrix |
| Bulnes et al. (2016) | [38] | Custom build | 95.5% | Genetic algorithm and artificial neural network |

## 3 Methodology

In this article, the NEU dataset was used with an 80:20 split ratio after augmentation as mentioned in Section 4.2 for pre-trained CNN models. The ImageNet weights were used for all CNN models by using the transfer learning technique. The validation result of each retained model is shown in Table 2 for comparison. Fig. 2 shows the flowchart of the study.

**Table 2:** Loss and accuracy of pre-trained CNN models concerning the batch size

| Model name | Batch size 128 | | Batch size 64 | | Batch size 32 | |
|---|---|---|---|---|---|---|
| | Val loss | Val accuracy | Val loss | Val accuracy | Val loss | Val accuracy |
| VGG16 | 0.1647 | 0.9520 | 0.1378 | 0.9533 | 0.0996 | 0.9653 |
| VGG19 | 0.2154 | 0.9332 | 0.1858 | 0.9325 | 0.1149 | 0.9628 |
| ResNet152 | 1.4654 | 0.5115 | 1.2296 | 0.5684 | 1.0664 | 0.6261 |
| ResNet152V2 | 0.2892 | 0.9496 | 0.3627 | **0.9583** | 0.4298 | 0.9543 |
| Xception | 0.1648 | 0.9613 | 0.2394 | 0.9514 | 0.2968 | 0.9506 |
| InceptionV3 | 0.2160 | 0.9525 | 0.2338 | **0.9586** | 0.2922 | 0.9570 |
| InceptionResNetV2 | 0.0938 | 0.9656 | 0.0774 | **0.9733** | 0.1462 | 0.9656 |
| NASNetLarge | 0.2362 | 0.9438 | 0.3008 | 0.9373 | 0.3885 | 0.9338 |
| MobileNetV2 | 0.1521 | 0.9689 | 0.1496 | **0.9755** | 0.5596 | 0.9568 |
| DenseNet201 | 0.0588 | 0.9826 | 0.0524 | **0.9837** | 0.1022 | 0.9791 |

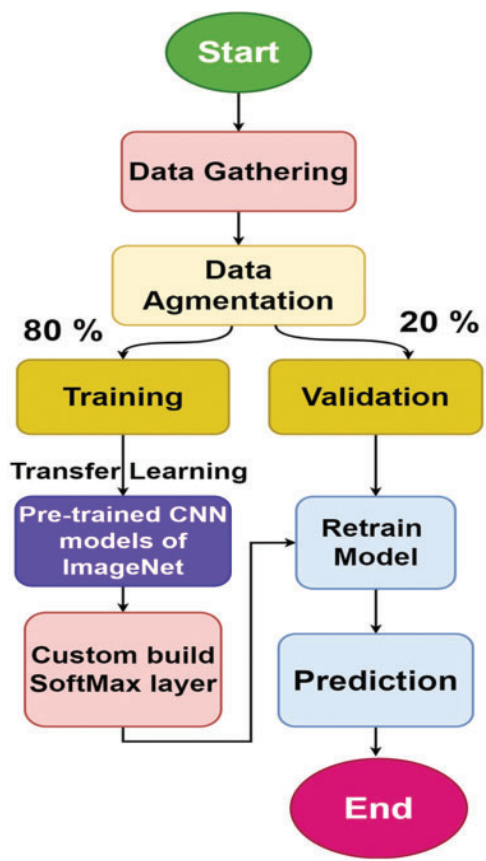

**Figure 2:** Flow chart of steel surface defect classification using transfer learning

### 3.1 Transfer Learning

Transfer learning is a novel technique that is also used in classification problems [39]. In transfer learning the knowledge, optimized weights, of a pre-trained network, which was trained on a large scale publicly available dataset, is transferred to another model for prediction same kind of other problems. Fig. 3 illustrates the overall process of transfer learning explaining each step in flow chart.

### 3.1.1 CNN Models

CNN is a kind of feed-forward neural network which consists of different convolutional and subsampling layers. The number of layers and size of each layer eventually depend on the complexity of the problem which is being evaluated. The CNN output for classification problems is the probability against each class. The motivation for selecting these models is based on their popularity and impact in the field of image classification. These models have been widely used in research and industry and have achieved state-of-the-art performance on benchmark datasets like ImageNet. Therefore, they serve as a good starting point for researchers who are interested in using pre-trained CNN models for steel defect detection. Additionally, each of these models has its unique architectural design and approach to optimizing performance, which can provide insights into the current trends and directions of CNN research.

### 3.1.2 Visual Geometry Group (VGG)

Simonyan [40] proposed small kernel sizes for deep CNN models, such as $3 \times 3$ convolutional filters, and named them VGG. The VGG was built on the simple principle of homogeneous topologies and trained on a large-scale dataset. The first VGG was 19 layers deep and achieved the second position in ILSVRC 2014. The main reason behind its popularity was its simplicity and homogeneous topologies. The main limitation of VGG19 was its 138 million parameters. A reduced version of VGG19 was VGG16 which is 16 layers deep.

### 3.1.3 ResNet

ResNet [16] introduced the concept of residual learning for deep networks. It supports multi-path-based CNNs. The ResNet proposed a 152-layer deep network and won the ILSVRC 2015. Moreover, ResNet also perform well and won first place in the COCO-2015 dataset [41].

### 3.1.4 Inception Networks

InceptionV3 and Inception-ResNet are the improved versions of Inception V1. The inceptionV3 reduces the computational cost by replacing the large-size filters of previous versions like $5 \times 5$ and $7 \times 7$ with small-size filters of $1 \times 5$ and $1 \times 7$. It also used $1 \times 1$ convolution before the large filters which exertion like a cross-channel correlation [42]. The Inception-ResNet was developed by combining the power of the residual network and inception block. Google's deep learning group introduces the inception block, a corresponding block, with the split, transform, and merge concept. It was observed that Inception-ResNet converges more quickly than Inception-V4 [43].

### 3.1.5 Xception Networks

Xception networks exploit the idea of depth-wise separable convolution [44]. Xception network replaces the spatial dimensions of the inception block of sizes $1 \times 1$, $5 \times 5$, and $3 \times 3$ with a single dimension of $3 \times 3$ followed by a $1 \times 1$ convolution. Xception network became more computationally efficient by decoupling spatial and feature map correlation.

### 3.1.6 NASNetLarge Network

NASNetLarge [45] is a convolutional neural network architecture developed by Google researchers in 2018. It consists of more than 88 million parameters and is deeper and wider than many popular CNN models like VGG16 and ResNet152. It uses a combination of normal and separable convolution layers, which reduces the computational complexity while maintaining a high level of accuracy.

### 3.1.7 MobileNet

Although CNN uses high computational costs, some of them are used for machine learning-based embedded systems like MobileNet. It is highly applicable for mobile and other resource-limited devices [42,43]. The basic unit of the MobileNet network is also depth-wise separable convolution. Which means the network perform a single convolution against each color channel rather than combining all color channel and flatting them. MobileNet is a 30-layer deep network.

### 3.1.8 DenseNet

The dense network has different variants like DenseNet121, DenseNet169, and DenseNet201. The DenseNet solves the vanishing gradient problem [46]. The vanishing gradient problem occurs when layers use more activation functions that are added to the neural network and eventually gradients of the loss function move toward zero and make the network hard to train. The DenseNet uses cross-layer connectivity in a modified fashion. Its passes all information from the previous layer to the next layer in a feed-forward fashion. So, the feature map of all preceding layers was used as inputs for all subsequent layers, as expressed in Eqs. (1) and (2). DenseNet concatenates the features rather than adding them. However, this approach makes the network parametrically expensive due to the increase in the number of feature maps at each layer [46].

The DenseNet-201 architecture consists of 4 dense blocks, each containing a series of convolutional layers with batch normalization and ReLU activation. The output of each dense block is passed through a transition layer, which reduces the spatial dimensions of the feature maps and increases the number of feature maps. The transition layer consists of a $1 \times 1$ convolutional layer, followed by a $2 \times 2$ average pooling layer. After the final dense block, a global average pooling layer is applied to the feature maps, followed by a fully connected layer with softmax activation to produce the final classification probabilities. Mathematically, the output of the kth layer in a DenseNet block can be written as:

$$X\_k = H\_k \left( [x\_0, x\_1, \ldots, x\_{k-1}] \right) \tag{1}$$

where x_i represents the output of the ith layer in the same block, and H_k is a composite function consisting of batch normalization, ReLU activation, and a $3 \times 3$ convolutional layer. The output of the final dense block is then passed through a transition layer, which can be written as:

$$x = \text{BN} \left( \text{ReLU} \left( \text{Conv} \left( x \right) \right) \right) * \text{pool\_size} \tag{2}$$

where BN represents batch normalization, ReLU is the rectified linear activation function, Conv is a $1 \times 1$ convolutional layer, and pool_size is the spatial downsampling factor in DenseNet-201. Finally, the output of the transition layer is passed through a global average pooling layer and a fully connected layer to produce the final classification probabilities using softmax activation. The softmax function computes the probability distribution y as:

$$y = \frac{ex_i}{\sum_{j=1}^{n} ex_j} \tag{3}$$

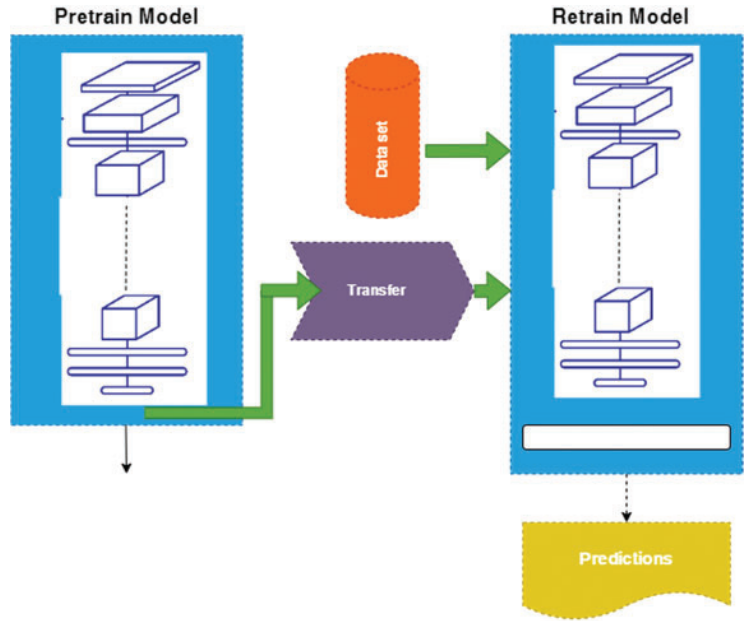

**Figure 3:** Transfer learning example for steel defect classification

## 4 Performance Analysis and Discussion

### 4.1 Dataset

Song et al. [47] of North Eastern University created the NEU Dataset. The NEU dataset is freely available to researchers, which encourages collaboration and knowledge sharing in the research community. This can lead to the development of new ideas and techniques for the analysis and processing of visual data in various fields. The NEU dataset provides a benchmark for researchers who are working on developing algorithms for the detection, classification, and recognition of metallic surfaces [48]. The NEU dataset is unique in that it includes images of various metallic surfaces under different lighting conditions, which makes it more realistic than other datasets [49]. As shown in Figs. 4 and 5, this dataset contains six different types of steel defects: rolled-in scale, patches, crazing, pitted surface, inclusion, and scratches. This dataset contains 1800 grayscale images with a size of $200 \times 200$ pixels. Each above-said class has 300 images, 240 (80%) images for training, and 60 images (20%) for testing. It is observed that:

— Inter-class steel defects have the same kind of physical characteristics.
— While intra-class reflects a clear and large difference in appearance [26].
— The illumination in the images becomes the cause of to change in the gray values of defected section and also becomes the possibility of noise [50] as shown in Fig. 4. The different data augmentation techniques, as mentioned in Section 4.2, were used for increasing the size of training data and making the model more robust and generalized.

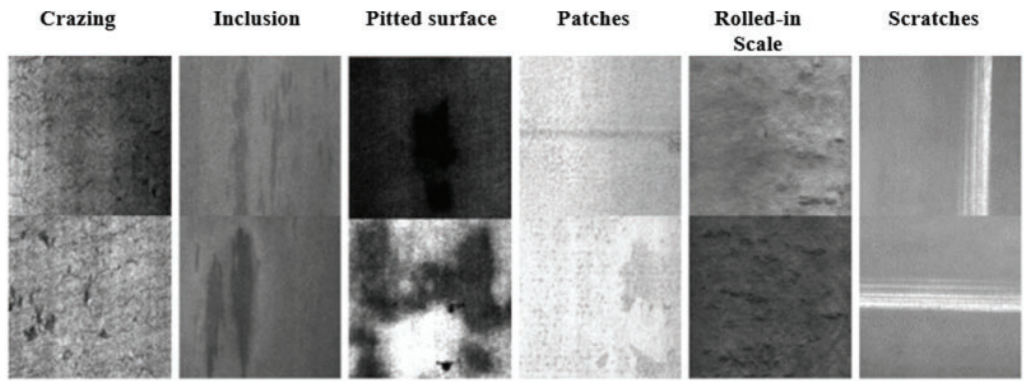

**Figure 4:** Six types of steel defects in the NEU dataset

### 4.2 Pre-Processing

Each class in the training dataset consists of only 240 images which are too small to train the CNN model as explained in Fig. 5. So different data augmentation techniques on training sets like vertical and horizontal flipping, rotation, height and width shift, and zooming are used to generate a large-scale unseen training dataset. The data augmentation techniques are not only used for artificially increasing the size of the data but also used for better generalization and avoiding over-fitting [6]. OpenCV (Open-Source Computer Vision) is a python library. Which was used for rescaling the images of steel in the experiment.

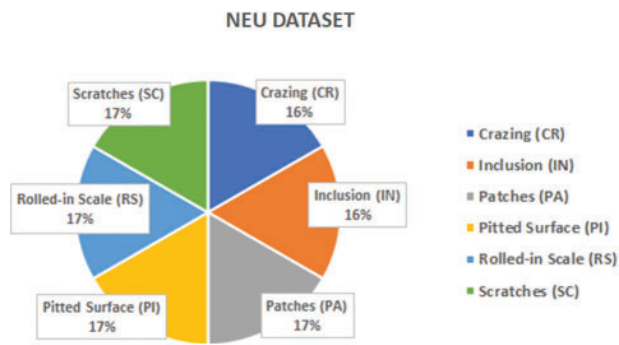

**Figure 5:** NEU dataset with its different classes and per class distribution

### 4.3 Experimental Design

The experiments are conducted by using python language because of its simplicity, which offers concise code and is suitable as a robust programing language. For developing the system, the KERAS API is used, which is supported by TensorFlow on the backend. 80% of the images of the dataset, as described in the above section, were used for training on NVIDIA Tesla T4 GPU and 20% of images were used for validation. The experiments were performed on different batch sizes like 128, 64, and 32 using Adam optimizer with a learning rate of 0.003. The pre-trained CNN models, as mentioned in Table 2, on the ImageNet dataset were used for performance evaluation. All the CNN models were fed with the same kind of data in respect of scaling, rotation, flipping, and zooming.

*4.4 Results Analysis*

The experiments of all the models are shown in Table 2. All the mentioned models were trained on the same kind of data as mentioned in Section 4.2 using the transfer learning technique with ImageNet weights with three different batch sizes: 128,64, and 32. The result of each model against each batch size was recorded separately.

The DenseNet201 shows a significant result and achieved 98.37 percent accuracy on the batch size of 64. The MobileNetV2 followed it with an accuracy of 97.55 percent on the same batch size of 64. Five models out of ten performed well on the batch size of 64 as compared to 128 and 32 batch sizes, highlighted values in Table 2. The ResNet152 shows the accuracy of 51.15, 56.84, and 62.61 on the batch size of 128, 64, and 32 respectively, which is significantly low among the others CNN models. The result also shows that ResNet152 performed well on small batch sizes rather than on large ones, Table 2.

In the confusion matrix of DenseNet201 on a batch size 64 Table 3, the rows represent the actual class labels, while the columns represent the predicted class labels. The abbreviations in the first row and column correspond to the different types of defects in the NEU dataset: Cr (crack), In (inclusion), Pa (patches), PS (pitted surface), RS (rolled-in scale), and Sc (scratches). For example, the cell at row 1, column 1 (Cr/Cr) shows that 93 instances truly belong to the "Cr" class and have been correctly classified as belonging to the "Cr" class. The remaining cell row 2, shows that have been incorrectly classified as belonging to the "Cr" class. In total, the model has achieved a high accuracy of 98.37% on the NEU surface defect dataset.

**Table 3:** The confusion matrix of DenseNet201 of batch size 64

|     | Cr | In | Pa | PS | RS | Sc |
|-----|----|----|----|----|----|----|
| Cr  | 93 | 1  | 2  | 1  | 0  | 3  |
| In  | 0  | 98 | 0  | 2  | 0  | 0  |
| Pa  | 2  | 0  | 97 | 1  | 0  | 0  |
| PS  | 0  | 1  | 0  | 97 | 0  | 2  |
| RS  | 1  | 0  | 0  | 0  | 98 | 1  |
| Sc  | 2  | 0  | 0  | 0  | 0  | 98 |

The DenseNet201 is a convolutional neural network (CNN) architecture that was introduced by Huang et al. in 2017. It consists of 201 layers, hence the name "DenseNet201" [51]. The architecture of the DenseNet201 model is based on the idea of dense connections between layers. In this architecture, each layer receives input not only from the previous layer but also from all the preceding layers. This creates a dense and efficient network that can learn features more effectively than traditional CNNs. The layered architecture is further explained in Fig. 6. The training accuracy and loss of DenseNet201 are shown in Fig. 8.

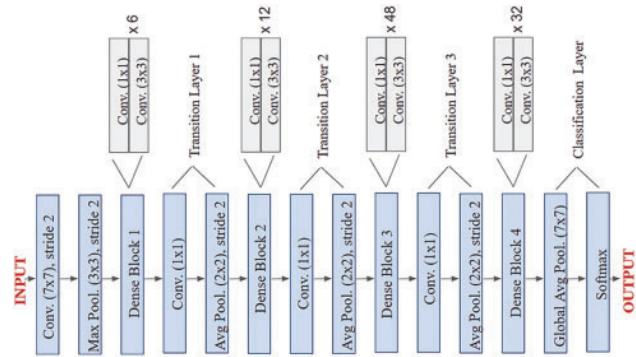

**Figure 6:** Layer architecture of DenseNet201

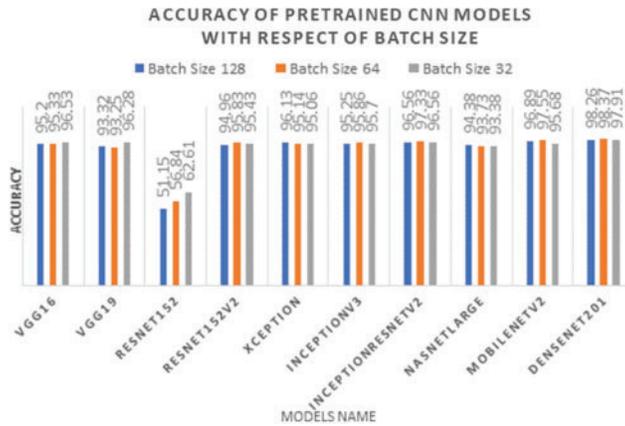

**Figure 7:** Accuracy of pre-trained CNN models with respect to batch size

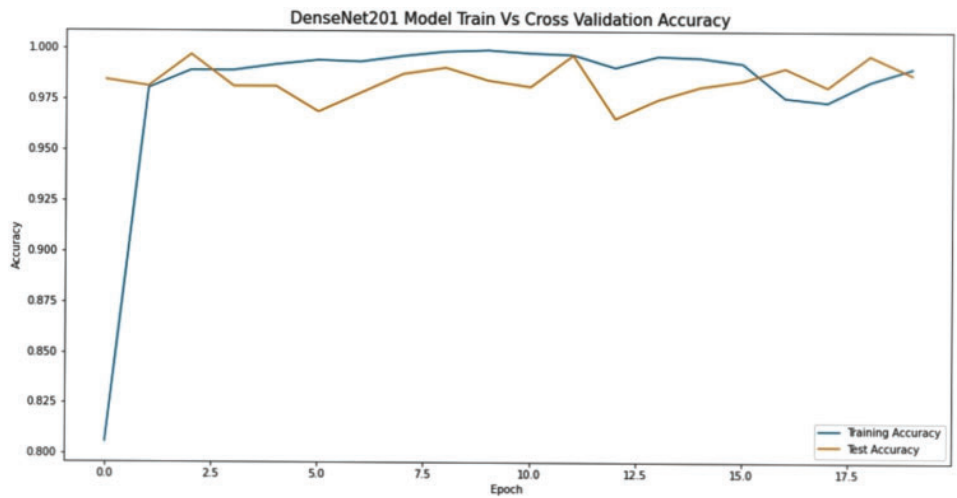

**Figure 8:** (Continued)

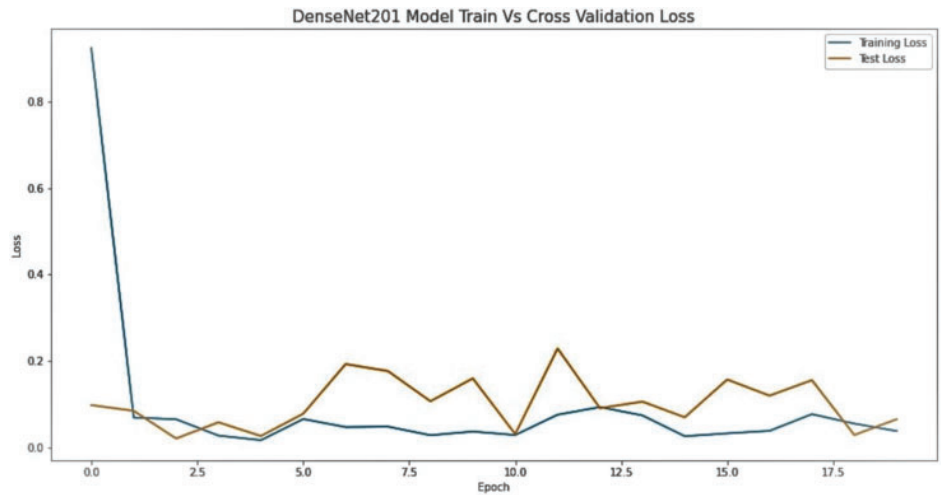

**Figure 8:** Training accuracy and loss of DenseNet201 on a batch size 64

## 5  Conclusion

The performance of different pre-trained CNN models for defect detection on the steel surface is being compared in this research article by using ImageNet weights on the NEU dataset with the split ratio of 80:20. The same kind of dataset for training all the models is used after applying different augmentation techniques as mentioned in Section 4.2. The models were trained using different batch sizes of 128, 64, and 32, as shown in Fig. 7. All the batches were used with the same learning rate of 0.003 and Adam optimizer. The results show that DenseNet201 performed significant results, with 98.37 percent accuracy on the batch size of 64. The training and loss graph of DenseNet201 is shown in Fig. 8. The MobileNetV2 likewise also produces remarkable results with an accuracy of 97.55 percent on the batch size of 64. These findings have important implications for the development of automated detection systems for steel surface defects, which can save time, reduce labor costs, and improve the quality of end products in various sectors. The study also contributes to the field of computer vision research by comparing the performance of different CNN models and providing guidance for researchers to select the most suitable model for similar tasks. Furthermore, future research can build on these findings by exploring the use of CNN models in other applications which are related to defect detection and quality control.

**Funding Statement:** This research work did not receive any specific funding from government, commercial, or non-profit organizations.

**Conflicts of Interest:** The authors declare that they have no conflicts of interest to report regarding the present study.